\documentclass[10pt,twocolumn,letterpaper]{article}
\pdfoutput=1
\newlength\minalignvsep

\usepackage{cvpr}
\usepackage{times}
\usepackage{epsfig}
\usepackage{graphicx}
\usepackage{amsmath}
\usepackage{amssymb}
\usepackage{pbox}
\usepackage{epstopdf}
\usepackage{subfigure}
\usepackage{xspace}
\usepackage{comment}
\usepackage{lipsum}
\usepackage{jeffe}
\usepackage{booktabs}
\usepackage{bm}
\usepackage{bbm}

\newcommand{\cmt}[2]{[#1: #2]}
\newcommand{\todo}[1]{\cmt{{\bf TODO}}{{\bf \color{blue} #1}}}

\newcommand{\myvec}[1]{\bm{#1}}

\newcommand{\shape}{S}

\newcommand{\graph}{\mathcal{G}}
\newcommand{\vertex}{\mathcal{V}}
\newcommand{\edge}{\mathcal{E}}

\newcommand{\mypara}{\vspace*{-15pt}\paragraph}
\newcommand{\myparaly}{\vspace*{-5pt}\paragraph}

\newcommand{\denselist}{\itemsep 0pt\parsep=0pt\partopsep 0pt\vspace{-2pt}}
\newcommand{\bitem}{\begin{itemize}\denselist}
\newcommand{\eitem}{\end{itemize}}
\newcommand{\benum}{\begin{enumerate}\denselist}
\newcommand{\eenum}{\end{enumerate}}
\newcommand{\bdescr}{\begin{description}\denselist}
\newcommand{\edescr}{\end{description}}

\cvprfinalcopy 

\def\cvprPaperID{832} 

\ifcvprfinal\fi
\begin{document}

\title{SyncSpecCNN: Synchronized Spectral CNN for 3D Shape Segmentation}

\author{Li Yi$^1$ \qquad Hao Su$^1$ \qquad Xingwen Guo$^2$ \qquad Leonidas Guibas$^1$\\$^1$Stanford University \qquad $^2$The University of Hong Kong}

\maketitle

\begin{abstract}
  In this paper, we study the problem of semantic annotation on 3D models that are represented as shape graphs.  A functional view is taken to represent localized information on graphs, so that annotations such as part segment or keypoint are nothing but 0-1 indicator vertex functions. Compared with images that are 2D grids, shape graphs are irregular and nonisomorphic data structures. To enable the prediction of vertex functions on them by convolutional neural networks, we resort to spectral CNN method that enables weight sharing by parameterizing kernels in the spectral domain spanned by graph laplacian eigenbases. Under this setting, our network, named SyncSpecCNN, strive to overcome two key challenges: how to share coefficients and conduct multi-scale analysis in different parts of the graph for a single shape, and how to share information across related but different shapes that may be represented by very different graphs. Towards these goals, we introduce a spectral parameterization of dilated convolutional kernels and a spectral transformer network. Experimentally we tested our SyncSpecCNN on various tasks, including 3D shape part segmentation and 3D keypoint prediction.  State-of-the-art performance has been achieved on all benchmark datasets.

\end{abstract}

\section{Introduction}
\label{sec:intro}
As has already happened in the image domain, the wide availability of 3D models brings with it the need to associate semantic information with the 3D data. In this work we focus on the problem of annotating 3D models represented by 2D meshes with part information. Understanding of the parts of an object (e.g., the back, seat and legs of a chair) is essential to its geometric structure, to its style, and to its function. 
There has been significant recent progress~\cite{Yi16} in the large scale part annotation of 3D models (e.g., for a subset of the ShapeNet~\cite{shapenet2015} models) -- our aim here is to leverage this rich data set so as to infer parts of new 3D object models. Our techniques can also be used to infer keypoints and other substructures within 3D models.

It is not straightforward to apply traditional deep learning approaches to 3D models because a mesh representation can be combinatorially irregular and does not permit the optimizations exploited by convolutional approaches, such as weight sharing, which depend on regular grid structures. In this paper we take a functional approach to represent information about shapes, starting with the observation that a shape part is itself nothing but a 0-1 indicator function defined on the shape. 

Our basic problem is to learn functions on shapes. We start with example functions provided on a given shape collection in the training set and build a neural net that can infer the same function when given a new 3D model. This suggests the use of a spectral formulation, based on a dual graph representing the shape, yielding bases for a function space built over the mesh.  

\begin{figure}
    \centering
    \includegraphics[width=0.4\textwidth]{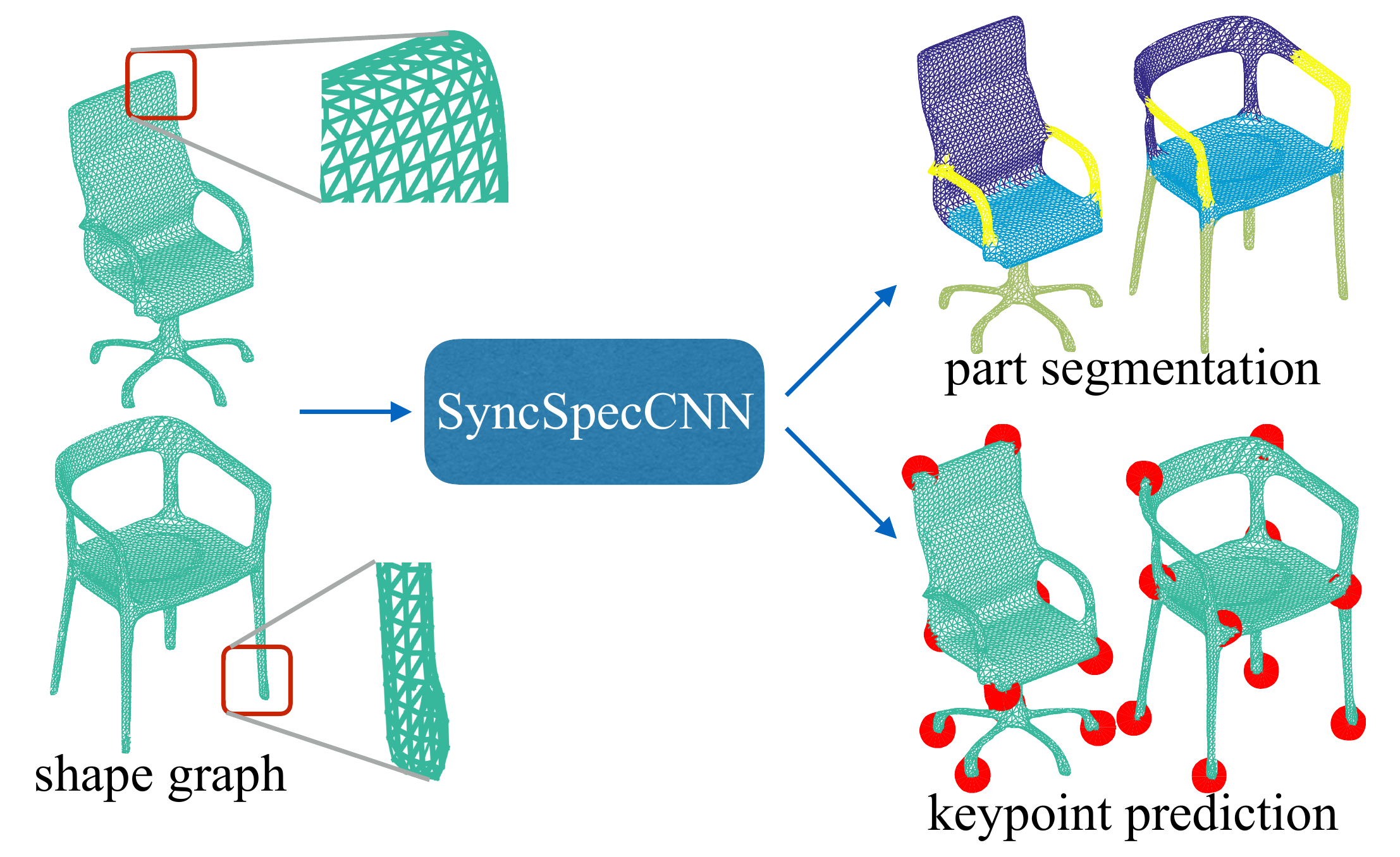}
    \caption{Our SyncSpecCNN takes a shape graph equipped with vertex functions (i.e. spatial coordinate function) as input and predicts a per-vertex label. The framework is general and not limited to a specific type of output. We show 3D part segmentation and 3D keypoint prediction as example outputs here. }
    \label{fig:teaser}
    \vspace{-0.3cm}

\end{figure}

With this graph representation we face multiple challenges in building a convolutional neural architecture. One is how to share coefficients and conduct multiscale analysis in different parts of the graph for a single shape. Another is how to share information across related but different shapes that may be represented by very different graphs. We introduce a novel architecture, the Synchronized Spectral CNN (SyncSpecCNN) to address these issues. 

The basic architecture of our neural network is similar to the fully convolutional segmentation network of~\cite{long2015fully}, namely, we repeat the operation of convolving a vertex function by kernels and applying a non-linear transformation. However, our network combines processing in the primal and dual (spectral) domains. 
We deal with the problem of weight sharing among convolution kernels at different scales in the primal domain by performing the convolutions in the spectral domain, where they become just pointwise multiplications by the kernel duals. 
Our key building block consists of passing to the dual, performing a pointwise multiplication and then returning to the primal representation in order to perform an appropriate non-linear step (such operations are not easily dualized).

The issue of information sharing across shapes is more challenging.  Since different shapes give rise to different nearest neighbor graphs on their point clouds, the eigenbases we get for the graph laplacians are not directly comparable. We synchronize all these laplacians by applying a functional map in the spectral domain to align them to a common canonical space. 
The aligning functional maps succeed in encoding all the dual information on a common set of basis functions where global learning takes place. An initial version of the aligning maps is computed directly from the geometry and then is further refined during training, in the style of a data-dependent spatial transformer network. 


We have tested our SyncSpecCNN on various tasks including 3D shape part segmentation and 3D keypoint prediction. We achieve state-of-the-art performance on all these tasks.

Key contributions of our approach are as follows:
\bitem
    \item We are the first to target at non-isometric shapes in the family of spectral CNNs.
    \item To allow weight sharing across different non-isometric shapes, we learn a Spectral Transformer Network.
    \item We introduce an effective spectral multiscale kernel construction scheme. 
\eitem

\section{Background}
\label{sec:related}
\myparaly{3D Shape Segmentation}
An important application of our framework is to obtain semantic part segmentation of 3D shapes in a supervised fashion. Along this track, most previous methods \cite{kalogerakis2010learning,xie20143d,makadia2014learning,guo20153d} employ traditional machine learning techniques and construct classifiers based on geometric features.
In the domain of unsupervised shape segmentation, there is one family of methods \cite{liu2004segmentation,liu2007mesh} emphasizes the effectiveness of spectral analysis for 3D shape segmentation. Inspired by this, our framework aims to marry the powerfulness of deep neural network and spectral analysis for 3D shapes segmentation.

\mypara{Spectral analysis on graphs}
We model a 3D shape $\shape$ as a graph $\graph=(\vertex,\edge)$, whose vertices $\vertex$ are points in $\R^3$ and edges $\edge$ connect nearby points. At each vertex of the graph, we can assign a vector. In this way, we define a vector-valued vertex function on $\graph$. For example, a segment on a shape can be represented as an indicator vertex function. this is the functional view of segmentation, as introduced in \cite{wang2013image,wang2014unsupervised}. The space of functions $\mathcal{F}$ defined on $\graph$ can be represented under different bases, i.e., $f=\sum_i \alpha_i \myvec{b}_i$ for $f\in\mathcal{F}$. One way to construct bases of $\mathcal{F}$ is through spectral analysis -- for each shape graph, the eigen vectors of its graph laplacian $L$ form an orthogonal bases $\myvec{B} = \{\myvec{b}_i\}$. One type of graph laplacian can be constructed as $L=I-D^{-1/2}WD^{-1/2}$, where $I$ is identity matrix, $D$ is the degree matrix and $W$ is the adjacency weight matrix of $\graph$. Under this construction, the eigenvalues $\myvec{\lambda}=\{\lambda_i\}$ corresponding to $\myvec{B}$ satisfy $0\le \lambda_i\le2$.

As is in Fourier analysis, the spectral decomposition also introduces the concept of frequency. For each basis $\myvec{b_i}$, the eigenvalue $\lambda_i$ in the decomposition defines its frequency, depicting its smoothness. By projecting $f$ on each basis $\myvec{b}_i$, the coefficient $\alpha_i$ can be obtained. $\myvec{\alpha} = \{\alpha_i\}$ is the spectral representation of $f$, in analogy to the Fourier transform. The convolution theorem of Fourier analysis can be extended to the laplacian spectrum: the convolution between a kernel and a function on the shape graph is equivalent to the point wise multiplication of their spectral representations \cite{bruna2013spectral,shuman2016vertex}.

\mypara{Functional map}
Different shapes define shape graphs with varied bases and spectral domains, which results in incomparable graph vertex function. Inspired by the recent work on synchronization \cite{singer2011angular,wang2013image,wang2014unsupervised}, we propose to align these different spectral domains using functional map \cite{ovsjanikov2012functional}. Functional map is initially introduced for this purpose on shapes. Specifically, given a pair of shape graph $\graph_i$ and $\graph_j$, a functional map from $\mathcal{F}_i$ to $\mathcal{F}_j$ is given by a matrix $X_{ij}$, which maps a function $f\in \mathcal{F}_i$ with coefficient vector $\myvec \alpha$ to the function $f'\in \mathcal{F}_j$  with coefficient vector $\myvec \alpha' = X_{ij} \myvec \alpha$. $\myvec \alpha$ and $\myvec \alpha'$ are computed according to a pair of bases. We refer the reader to \cite{ovsjanikov2012functional} for detailed introduction and intuition.  

\mypara{CNN on Graphs}
We call such CNNs as ``graph CNNs''. Graph CNNs takes a graph with vertex function as input. Conventional image CNN can be viewed as a graph CNN on 2D regular grids of pixels, with RGB values as the vertex function. There have been some previous work studying graph CNN on more general graphs instead of 2D regular grids \cite{bruna2013spectral,duvenaud2015convolutional,henaff2015deep,defferrard2016convolutional}, and \cite{masci2015geodesic,boscaini2015learning,boscaini2016learning} have a special focus on near-isometric 3D shape graphs like human bodies. To generalize image CNN, These work usually tries to tackle the following three challenges: defining translation structures on graphs to allow parameter sharing; designing compactly supported filters on graphs; aggregating multi-scale information. Their constructions of deep neural network usually fall into two types: spatial construction and spectral construction. The approach we propose belongs to the family of spectral construction but with two key differences: we explicitly design an effective multi-scale information aggregation scheme; we synchronize different spectral domains to allow parameter sharing among very different shape graphs thus increasing generalizability of our SyncSpecCNN.

\section{Problem}
\label{sec:problem}
Given a 3D shape $\shape$ represented as a shape graph $\graph=(\vertex,\edge)$, we seek for a per-vertex label $l$, such as segmentation or keypoints. These labels are represented as vertex functions $f$ on $\graph$, i.e., $f:\vertex\rightarrow \R^K$. We precompute a set of 3D features for each vertex $v \in \vertex$ and use them as input vertex functions. These features capture location, curvature, and local context properties of each vertex $v$ and we use the publicly available implementation \cite{kim2014shape2pose}. To represent the functional space on shape graphs $\graph$, we also construct the graph laplacian $L$ of each shape $\shape$, compute the spectral frequency $\myvec{\lambda}=\{\lambda_i\}$ and corresponding bases $\myvec{B}=\{\myvec{b}_i\}$ through eigendecomposition. We note that a basis $\myvec{b}_i$ is also a vertex function. Therefore, our neural network takes the laplacian $L$ of a graph $\graph$ and vertex functions of local geometric features as input, and predicts a vertex function $f$ such as segmentation or keypoint indicator function.

\section{Approach}
\subsection{Overview}
The basic architecture of our SyncSpecCNN is similar to the fully convolutional segmentation network as in \cite{long2015fully}, namely, we repeat the operation of convolving the vertex function by kernels and applying non-linear transformation. However, we have several key differences. First, we achieve convolution by modulation in the spectral domain.
Second, we parametrize kernels in the spectral domain following a dilated fashion, so that kernel sizes could be effectively enlarged to capture large context information without increasing the number of parameters.
Last, we design a Spectral Transformer Network to synchronize the spectral domain of different shapes, allowing better parameter sharing.

\subsection{Network Architecture}
\begin{figure*}
    \centering
    \includegraphics[width=0.9\linewidth]{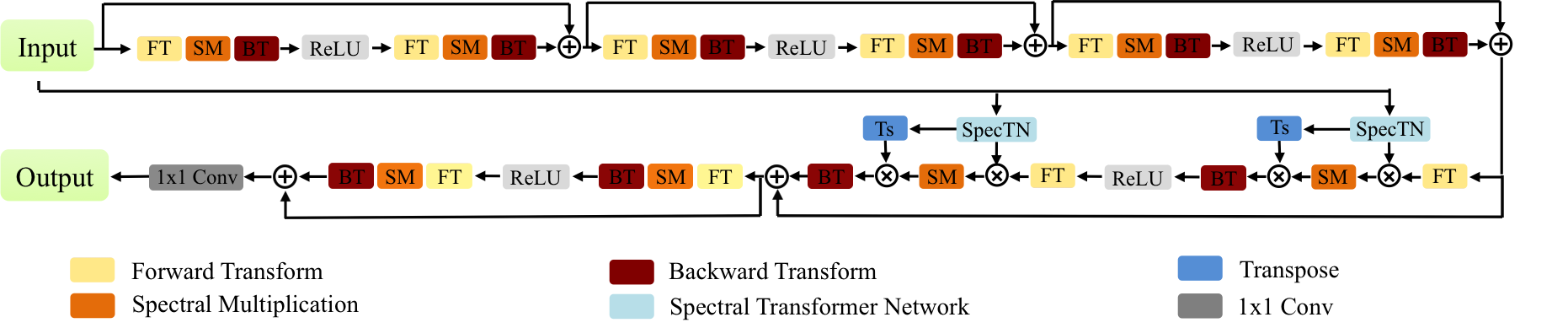}
    \caption{Architecture of our SyncSpecCNN. Spectral convolution is done through first transforming graph vertex functions into their spectral representation and then pointwise modulating it with a set of multipliers. The multiplied signal is transformed back to spatial domain to perform nonlinear operations. We introduce spectral transformer network to synchronize different spectral domains and allow better parameter sharing in spectral convolution. Convolution kernels are parametrized in a dilated fashion for effective multi-scale information aggregation.}
    \label{fig:architecture}
\end{figure*}

Similar to conventional CNN, our SyncSpecCNN contains layers including  ReLU, DropOut, 1$\times$1 Convolution \cite{szegedy2015going}, and BatchNormalization, which all operate in the spatial domain on graph vertex functions. The difference comes from our graph convolution operation, which introduces the following modules: Forward Transform, Backward Transform, Spectral Multiplication, and Spectral Transformer Network, as is shown in Figure~\ref{fig:architecture} and summarized in Table~\ref{tab:architecture}. 

We provide more details about the newly introduces modules as below.

\begin{table}[]
\centering
{\footnotesize
\begin{tabular}{@{}p{0.25\linewidth}p{0.015\linewidth}p{0.015\linewidth}p{0.015\linewidth}p{0.015\linewidth}p{0.015\linewidth}p{0.015\linewidth}p{0.017\linewidth}p{0.016\linewidth}p{0.016\linewidth}p{0.015\linewidth}}
\toprule
Layer               & 1  & 2  & 3  & 4  & 5  & 6  & 7   & 8   & 9  & 10 \\ \midrule
Dilation ($\gamma$) & 1  & 1  & 4  & 4  & 16 & 16 & 64   & 64   & 1  & 1  \\
SpecTN              & No & No & No & No & No & No & Yes & Yes & No & No \\
\#Kernel Param      & 7  & 1  & 7  & 1  & 7  & 1  & 45  & 45  & 7  & 1  \\
\#Out Channel    & c  & c  & c  & c  & 2c & 2c & 2c  & 2c  & 2c & 2c\\ \bottomrule
\end{tabular}
}
\caption{Parameters used in different layers of the architecture, including dilation parameter $\gamma$ which controls convolution kernel size, whether use spectral transformer network (SpecTN), the number of learnable parameters in convolution kernels, the number of output channels after each convolution operation.}
\label{tab:architecture}
\end{table}

In a basic convolution block, a vertex function $f$ defined on $\graph$ is first transformed into its spectral representation $\myvec{\alpha}$ through Forward Transform $\myvec{\alpha} = \myvec{B}^Tf$. Then the functional map $C$ predicted by the Spectral Transformer Network will be applied to $\myvec{\alpha}$ and outputs $\myvec{\alpha}'=C\myvec{\alpha}$ for spectral domain synchronization (Sec~\ref{spectn}). A Spectral Multiplication layer is followed, pointwisely multiplying $\myvec \alpha'$ by a set of multipliers and getting $\tilde{\myvec \alpha}' = W\myvec \alpha'$, where $W$ is a diagonal matrix with its diagonal being the set of multipliers, and $\tilde{\myvec \alpha}'$ is used to denote the multiplication result. This is how we conduct convolution in the spectral domain, where spectral dilated kernels are used to capture multiscale information (Sec~\ref{sdkp}). Then we apply the inverse functional map $C_{inv}$ to $\tilde{\myvec \alpha}'$, so that we get the spectral representation $\tilde{\myvec \alpha} = C_{inv}\tilde{\myvec \alpha}'$ in the original spectral domain before canonicalization. $\tilde{\myvec \alpha}$ is then converted back to a graph vertex function through Backward Transform $\myvec \tilde{f} = \myvec{B}\tilde{\myvec \alpha}$. This building block was repeated for several times and forms the backbone of our deep architecture. We also add skip links into our SyncSpecCNN to better facilitate information flow across earlier and later layers. 

One interesting observation is worth mentioning: small convolution kernels correspond to smoothly transiting multipliers in the spectral domain, therefore not very sensitive to bases misalignment among shapes graphs in a certain range of spectrum and are more generalizable across graphs. As a result, we omit the spectral transformer network when the convolution kernels are small. 

\subsection{Spectral Dilated Kernel Parameterization}
\label{sdkp}
Yu et al.~\cite{yu2015multi} has proved the effectiveness of multi-scale kernels for aggregating context information at different scales in the task of image segmentation. They propose to use dilated kernels to increase the kernel size without increasing the number of parameters. We parametrize our convolution kernels in a similar flavor but in the spectral domain, which turns out to be straightforward and effective. Essentially, we find that multi-resolution analysis on graphs could be achieved without complicated hierarchical graph clustering.

Before explaining what the exact parametrization is, we first discuss the intuition behind our design. The Spectral Multiplication layer modulates the spectral representation $\myvec \alpha=\{\alpha_i\}$ by a set of multipliers from the kernel, where $\alpha_i$ is the spectral coordinate of vertex function at basis $\myvec{b}_i$. Note that $\lambda_i$ can be interpreted as the frequency of its corresponding eigenbasis $\myvec b_i$, and $\myvec b_i$ itself is a vertex function that captures the intrinsic geometry of the shape. We assume that $\lambda_i$'s are sorted ascendingly and arrange $\myvec{b}_i$'s accordingly.

The multiplers are the spectral representation of convolution kernel. Denote the set of multipliers as $\myvec m=\{m_i\}$, each corresponds to one $\lambda_i$. Regard $\myvec m$ as a function of $\lambda_i$. 

Again, generalized from conventional Fourier analysis, if $\myvec m$ is concentrated in the low-end of the spectrum, the corresponding spatial kernel function is smooth; conversely, if the corresponding spatial functions is localized, $\myvec m$ is smooth. Therefore, to obtain a smoother kernel function as in \cite{yu2015multi}, we constrain the bandwidth of $\myvec m$, enabling us to learn a smaller number of parameters; in addition, varying the smoothness of $\myvec m$ would control the kernel size. 

To be specific, we associate each Spectral Multiplication layer with a dilation parameter $\gamma$ and parameterize $m_i$ as a combination of some modulated exponential window functions, namely

\vspace{-0.25cm}
\begin{align*}
    m_i = \sum_{j=0}^n\omega_{2j+1}\text{e}^{-j\gamma \lambda_i}\text{cos}(j\gamma \lambda_i \pi)
    +\sum_{j=1}^n\omega_{2j}\text{e}^{-j\gamma \lambda_i}\text{sin}(j\gamma \lambda_i \pi)
    \vspace{-0.25cm}
\end{align*}

\begin{figure}
    \centering
    \includegraphics[width=0.8\linewidth]{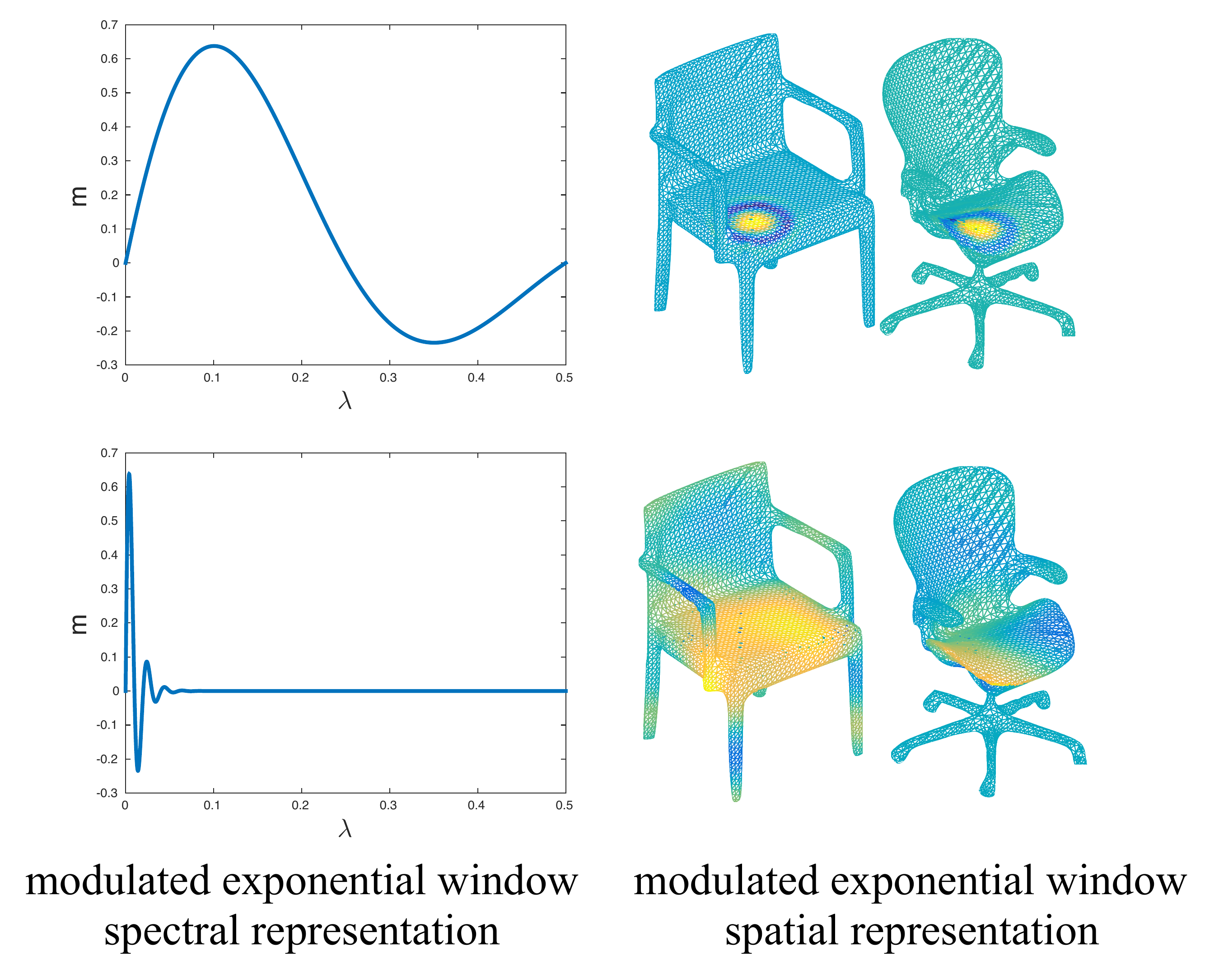}
    \caption{Visualization of modulated exponential window function with different dilation parameters in both spectral domain and spatial domain. The same spectral representation could induce spatially different kernel functions, especially when the kernel size is large.}
    \label{fig:kernelvis}
    \vspace{-0.3cm}
\end{figure}

Here $\myvec \omega$ is a set of $2n+1$ learnable parameters, $n$ is a hyper-parameter controlling the number of learnable parameters. Large $\gamma$ corresponds to rapidly changing multipliers with small bandwidth, thus a smooth kernel with large spatial support. On the other hand, small $\gamma$ corresponds to slowly changing multipliers with large bandwidth, corresponding to kernels with small spatial support. Instead of using an exponential window only, we add $\text{sin}/\text{cos}$ modulation to increase the expressive power of the kernel. Figure~\ref{fig:kernelvis} shows a visualization of modulated exponential window function with different dilation parameter.

Our parametrization has three main advantages: First, it allows aggregating multi-scale information since the size of convolution kernels vary in different layers; Second, large kernels could be easily acquired with a compact set of parameters, which effectively increases the receptive field while mitigates overfitting; Third, reduced parameters allow more efficient computation.

\vspace{-0.1cm}

\subsection{Spectral Transformer Network}
\label{spectn}
As is shown in Figure~\ref{fig:kernelvis}, the same spectral parametrization of kernels could lead to very different vertex functions when the underlying spectral domains are different. This problem is especially prominent when the kernel size is large. Therefore, being able to synchronize different spectral domains is the key to allow large kernels sharing parameters across different shape graphs. 

\subsubsection{Basic idea} According to \cite{ovsjanikov2012functional} and \cite{wang2013image}, one way to synchronize the spectral domains of a group of shapes is through a tool named functional map. In the functional map framework, one can find a linear map to pull the spectral domain of each individual shape to a canonical space, so that representations in the individual spectral domains become comparable under a canonical set of bases. Indeed, given each shape $S$, this linear map is as simple as a matrix $C$, which linearly transforms the spectral representation $\myvec \alpha$ on one shape to its counterpart $\myvec \alpha'$ in the canonical space. Note that, {\bf from the synchronization in the spectral domain, one induces a spatial correspondence on the graph, vice versa}. Viewing the spectral domain as the dual space and spatial domain on graph as the primal space, this {\bf primal-dual relationship} is the pivotal idea behind functional map.

\begin{figure}[t!]
    \centering
    \includegraphics[width=0.4\textwidth]{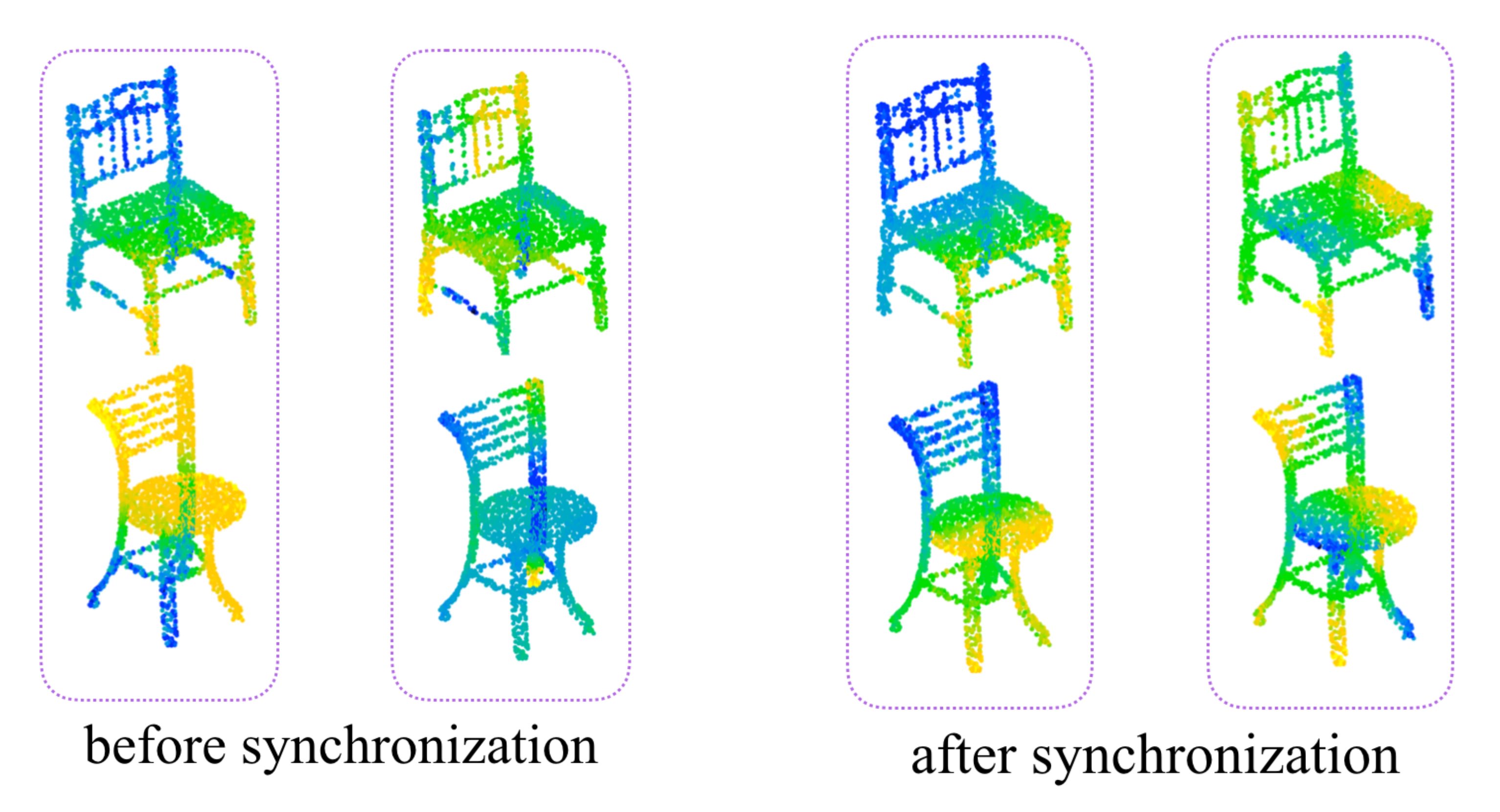}
    \caption{Visualization of low frequency eigenbasis functions before and after spectral synchronization. Before synchronization, eigenbasis functions on different shapes are not aligned. After applying the transform predicted from SpecTN, different spectral domain could be synchronized and the eigenbasis functions align.}
    \label{fig:jointbasis}
\end{figure}

Inspired by this idea, we design a Spectral Transformer Network (SpecTN) for the spectral domain synchronization task. Our SpecTN takes a shape $S$ as input and predicts a matrix $C$ for it (see Figure~\ref{fig:architecture}), so that
$\myvec \alpha'=C \myvec \alpha$.
Thus, without SpecTN, $\myvec \alpha$ will be directly passed to subsequent modules of our network; with SpecTN, $\myvec \alpha'$ will be passed. In Figure~\ref{fig:jointbasis}, we show an example of how different spectral domains are synchronized after applying the linear map $C$ predicted from our SpecTN.

Our SpecTN draws inspiration from Spatial Transformer Network (STN)~\cite{jaderberg2015spatial}. From a high level, both SpecTN and STN are learned to align data to a canonical form.

\subsubsection{Input to SpecTN}
A proper representation for shape $\shape$ is needed as the input to our SpecTN. To allow SpecTN predicting a transform between different spectral domains, certain depiction about the underlying spectral domain is greatly helpful, i.e. graph laplacian eigenbases in our setting. In addition, since spectral synchronization couples with graph alignment, providing rough shape graph correspondences could facilitate good prediction. 

Based on these, we use voxel functions $\myvec{B}_v$ that is computed from laplacian eigenbases as the input to SpecTN:
    $C=\mbox{SpecTN}(\myvec{B}_{v};\Theta).$
Specifically, $\myvec{B}_v$ is a \emph{volumetric reparameterization} of the graph laplacian eigenbases $\myvec{B}$, defined voxel-wise in 3D volumetric space. The volumetric reparameterization is conducted by converting graph vertex function $\myvec{B}$ into voxel function $\myvec{B}_v$ in a straightforward manner -- we simply assign a vertex function value to the voxel where the vertex lies. Since all $\myvec{B}_v$ live in the same 3D volumetric space, correspondences among them are associated accordingly.


\subsubsection{Optimization of SpecTN}
Ideally, SpecTN should be learned automatically along with the minimization of the prediction loss, as the case in STN; however, in practice we find that such optimization is extremely challenging. This is because the parameters of $C$ in SpecTN is quadratic w.r.t the number of spectral bases, hundreds of times more than in the affine transformation matrix of STN. 

We address this challenge from three aspects: limit our scope to a reduced set of prominent spectral bases to curtail the parameters of $C$; add regularization to constrain the optimization space; smartly initialize SpecTN with a good starting point.

\mypara{Reduced bases} Synchronizing the whole spectrum could be a daunting task given its high dimensionality. In particular, free parameters in $C$ grows quadratically as the dimension of spectral domain increases. To favor optimization, we adopt a natural strategy that only synchronizes the prominent part of the spectrum. In our case, the spectral parametrization of large kernels are mainly determined by the low-frequency end of the spectrum, indicating that the synchronization in this part of spectrum is sufficient. In practice, we synchronize the top $15$ bases sorted by the frequency. This idea has been verified to be effective by \cite{ovsjanikov2012functional}.



\mypara{Regularization}
Regularizations are used during training to force the output $C$ of SpecTN to be close to an orthogonal map, namely, in the overall loss function we add a term $\|CC^T-I\|_F^2$. With this regularization, $C^T$ can be used to approximate the inverse map. Such a maneuver is more friendly to differentiation and easier to train.

\mypara{Initialization by precomputed functional map} 
Given the huge optimization space and the non-convex objective, a good starting point helps to avoid optimization from getting stuck in bad local minima. As stated above, our linear transformation $C$ can be interpreted as a functional map; therefore, it is natural for us to initialize $C$ accordingly and then refine it to better serve the end-task. To this end, we first precompute a set of function maps $C_{pre}$ for each shape by an external routine, which roughly align each individual spectral domain of $\shape$ to a canonical domain. Then we pretrain the SpecTN separately in a supervised manner:
\begin{equation*}
\begin{aligned}
\underset{\Theta}{\mbox{minimize}} & &\sum_i \|\mbox{SpecTN}(\myvec{B}_{v, i};\Theta) - C_{pre, i}\|^2
\end{aligned}
\end{equation*}
where $i$ indexes shapes.
This pretrained SpecTN is plugged into the  SyncSpecCNN pipeline and fine-tuned while optimizing a specific task such as shape segmentation.  Validated by our experiment, the pretraining step is crucial. 

Next we introduce how the external routine precomputes a functional map for some shape $\shape$. This functional map aligns the spectral domain of $S$ to a canonical one of an ``average'' shape $\bar{\shape}$. So we start from the construction of the ``average'' shape and then proceed to the computation of the functional map. 

The geometry of $\bar{\shape}$ is not generated explicitly. Instead, $\bar{\shape}$ is represented by its volumetric adjacency matrix $\bar{W}_v$, which depicts the connectivity of voxels in the volumetric space that all shapes are voxelized. $\bar{W}_v$ is obtained by averaging the volumetric adjacency matrices $W_v$ of all shapes. The $W_v$ for each shape $S$ is the adjacency matrix of the corresponding volumetric graph, whose vertices are all the voxels and edges indicate the adjacency of occupied voxels in the volumetric space.

The functional map $C$ from $\shape$ to $\bar{\shape}$ could be induced from the spatial correspondences between $\shape$ and $\bar{\shape}$, by the primal-dual relationship~\cite{ovsjanikov2012functional}. Since we already have the bases of $\shape$ and $\bar{\shape}$, as well as the rough spatial correspondences between them from the volumetric occupancy, this map can then be discovered by the approach proposed in \cite{ovsjanikov2012functional}. To be specific, we use $\myvec{B}_v$ to denote the volumetric reparametrization of graph laplacian eigenbases $\myvec{B}$ for each shape $\shape$, and use $\bar{\myvec{B}}_v$ to denote the grahp laplacian eigenbases of $\bar{\shape}$. $\myvec{B}_v$ and $\bar{\myvec{B}}_v$ both lie in the volumetric space and their spatial correspondence is natural to acquire. The functional map $C_{pre}$ aligning $\myvec{B_v}$ with $\bar{\myvec{B}}_v$ could be computed through simple matrix multiplication $C_{pre}=\bar{\myvec{B}}_v^T\myvec{B}_v$. The computed functional map will serve as supervision and SpecTN is pretrained to minimize the loss function $||C-C_{pre}||_F^2$.

It is worth mentioning that, if the shapes under consideration are diverse in topology and geometry, i.e. shapes from different categories, aligning every shape to a single ``average'' shape might cause unwanted distortion. Therefore we leverage multiple ``average'' shapes $\{\bar{\shape}_i\}_{i=1}^{n}$ and use a combination of their spectral domains as the canonical domain. Specifically, we assign each shape $\shape$ to its closest ``average'' shape under some global similarity measurement (i.e. lightfield descriptor) and use $\{a_i\}_{i=1}^n$ to represent such assignment, namely $a_i=1$ if $\shape$ is assigned to $\bar{\shape}_i$ and $a_i=0$ otherwise. Also we use $\bar{\myvec{B}}_{vi}$ to denote the spectral bases of $\bar{\shape}_i$. Then the functional map $C_{pre}$ for each shape $\shape$ could be computed through $C_{pre}=[a_1\bar{\myvec{B}}_{v1} \;a_2\bar{\myvec{B}}_{v2}\; ...\; a_n\bar{\myvec{B}}_{vn}]^T\myvec{B}_v$. The SpecTN is pretrained to predict a functional map which only synchronizes spectral domain of each shape to its most similar ``average'' shape.

\subsection{Implementation Details}
\label{sec:impl}
In most of our experiments, input shapes are represented as point cloud with around $2000-3000$ points. Given an input shape point cloud, we build a k-nearest neighbor graph $\graph$ first. We use $k=6$ in all our experiments. Then a graph weight matrix $W$ could be constructed in which $W_{i,j}=\frac{1}{d_{i,j}^2}$ if point $i$ and $j$ are connected, $0$ otherwise. We then compute the symmetric normalized graph laplacian $L$ as $L=I-D^{-1/2}WD^{-1/2}$, where $D$ is the degree matrix and $I$ denotes identity matrix. Since many natural functions we care about could be depicted by a small number of low-frequency laplacian eigenbases, we compute and use the smallest $100$ eigenvalues as well as the corresponding eigenbases for each $L$ in all our experiments. 

The choice of dilation parameters $\gamma$, number of output channels after each convolution layer, number of learnable parameters in each convolution kernel are shown in Table~\ref{tab:architecture}. We choose $c=50$ in all of our experiments. As is mentioned, we only consider the problem of synchronizing the low-frequency end of different spectral domains, so we choose to predict a functional map $C\in\mathbb{R}^{15\times45}$ in our experiments, which maps the first $15$ eigenbases of each individual spectral domain into a canonical domain of dimension $45$. Notice the dimension of canonical domain is larger that each individual domain to allow very different shapes to be mapped into different subspaces.

\section{Experiment}
\begin{table*}[t!]
\centering
\small
\begin{tabular}{@{}p{0.09\linewidth}|p{0.04\linewidth}|p{0.025\linewidth}p{0.025\linewidth}p{0.025\linewidth}p{0.025\linewidth}p{0.03\linewidth}p{0.025\linewidth}p{0.025\linewidth}p{0.025\linewidth}p{0.03\linewidth}p{0.03\linewidth}p{0.025\linewidth}p{0.03\linewidth}p{0.03\linewidth}p{0.03\linewidth}p{0.03\linewidth}p{0.03\linewidth}}
\hline
category & mean & plane & bag & cap & car & chair & ear-phone & guitar & knife & lamp & laptop & motor-bike & mug & pistol & rocket & skate-board & table \\ \hline
Wu14 \cite{wu2014interactive} & - & 63.20 & - & - & - & 73.47 & - & - & - & 74.42 & - & - & - & - & - & - & 74.76 \\
Yi16 \cite{Yi16} & 81.43 & 80.96 & 78.37 & 77.68 & \textbf{75.67} & 87.64 & 61.89 & 91.79 & 85.36 & 80.59 & 95.58 & \textbf{70.59} & 91.85 & \textbf{85.94} & 53.13 & 69.81 & 75.33 \\
ACNN \cite{boscaini2016learning} & 79.63 & 76.35 & 72.89 & 70.80 & 72.72 & 86.12 & 71.14 & 87.84 & 81.98 & 77.43 & 95.49 & 45.68 & 89.49 & 77.41 & 49.23 & 82.05 & 76.71 \\
Voxel CNN & 79.37 & 75.14 & 72.80 & 73.28 & 70.00 & 87.17 & 63.50 & 88.35 & 79.58 & 74.43 & 93.92 & 58.67 & 91.79 & 76.41 & 51.16 & 65.25 & 77.08   \\ \hline
Ours1 & 83.48 & 80.61 & 81.62 & 76.92 & 73.86 & 88.65 & 74.48 & 89.03 & 85.34 & 83.47 & 95.53 & 62.74 & 92.01 & 80.88 & \textbf{62.10} & 82.23 & 81.36 \\
Ours2 & \textbf{84.74} & \textbf{81.55} & \textbf{81.74} & \textbf{81.94} & 75.16 & \textbf{90.24} & \textbf{74.88} & \textbf{92.97} & \textbf{86.10} & \textbf{84.65} & \textbf{95.61} & 66.66 & \textbf{92.73} & 81.61 & 60.61 & \textbf{82.86} & \textbf{82.13} \\ \hline
\end{tabular}
\caption{IoU for part segmentation on 16 categories. To compute mean IoU, per category IoU is weighted by the corresponding shape number and then averaged. Ours1 represents a variation of our framework without SpecTN and Ours2 corresponds to our full pipeline with SpecTN. On average, our approach outperforms all the baseline including both traditional machine learning and deep learning based methods by a large margin. We also achieves the highest IoU on most of the categories.}
\label{tab:percatseg}
\end{table*}
\label{sec:exp}
Our proposed SyncSpecCNN takes one graph vertex function as input and predicts another as output. As a generic framework, the prediction is not limited to a specific type of graph vertex function and can be tailored towards different goals. To evaluate the effectiveness of our framework, we divide our experiments into five parts. First, we evaluate on a benchmark of 3D shape segmentation~\cite{shapenet2015,Yi16}. Second, we evaluate on keypoint prediction task using a new large scale keypoint annotation dataset. Third, we leverage SyncSpecCNN to learn vertex normal functions and visualize the prediction results qualitatively. Fourth, we perform control experiments to compare different design choices of the framework and analyze the stability of our system under input sampling density variations. Last, we show qualitative results and analyze error patterns.

\subsection{Dataset}
For 3D shape segmentation task, we use a large scale shape part annotation dataset introduced by \cite{Yi16}, which augments a subset of ShapeNet models with semantic part annotations. The dataset contains 16 categories of man-made shapes, with 2 to 6 parts per category. In total there are 16,881 models with expert verified part annotations. In addition, we use the official train/test split provided along with ShapeNet models.

For the keypoint prediction task, we build a new large scale keypoint annotation dataset, containing 1,337 chair models with 10 keypoints per shape, in contrast to traditional small scale dataset \cite{kim2013learning} which has at most 100 shapes annotated per category. These keypoints are all manually annotated by experts with consistency across different shapes. 

\subsection{Shape Part Segmentation} 

\myparaly{Per-category shape part segmentation}
We first conduct part segmentation assuming the category label of each shape is known, as the setting in \cite{Yi16}. The task is to predict a part label for each sample point on shapes. We compare our framework with traditional learning-based techniques \cite{wu2014interactive,Yi16} leveraging on local geometric features and shape alignment cues, as well as recent deep learning based approaches \cite{boscaini2016learning} which also fall into the family of spectral CNNs. In addition we design an additional baseline using a 3D volumetric CNN architecture, denoted as Voxel CNN, which generalizes VoxNet~\cite{maturana2015voxnet} for segmentation tasks. The network has 10 convolutional layers without down-sampling and keeps a receptive field of 19 with spatial resolution of 32. We compute per-point features in the preprocessing step as is in \cite{Yi16} and use the same set of input for all baselines except Voxel CNN. The set of input shapes are pre-aligned using a hierarchical joint alignment algorithm described in \cite{shapenet2015}. Point intersection over union (IoU) is used as evaluation metric, averaged across all part classes. Cross-entropy loss is minimized during training. 

We evaluate our framework in two settings, with or without SpecTN, and compare the results in Table~\ref{tab:percatseg}. 

Note that on most categories our approach achieves the best performance and on average outperforms state of the art by a large margin. In comparison to \cite{boscaini2016learning},  the state of the art in the family of spectral CNNs, our approach introduces spectral dilated kernel parametrization, which increases the effectiveness of spectral CNN framework.  Moreover, the performance gain from SpecTN shows that synchronizing spectral domains would greatly increase the generalizibility across shapes of different topology and geometry.

\mypara{Cross-category shape part segmentation}
Next we evaluate our approach on the part segmentation task in a cross-category setting. In this task, shape category label is not known during the test phase and for each point the network needs to select one of the part label from all possible part labels in all categories. Cross-category setting introduces larger geometric and topological variance among shapes, thus could help examining the spectral CNN's ability of recognizing objects. At the same time the impact of spectral domain misalignment becomes stronger, providing a better testbed for validating the effectiveness of SpecTN. 
Since this experiment is proposed to verify design choices of spectral CNN, we mainly compare with \cite{boscaini2016learning}. We mix the 16 categories of shapes in \cite{Yi16} and train a single network for all categories. After predicting point segmentation labels, one can classify shapes through a point-wise majority voting scheme. Point IoU and classification accuracy (Acc) are chosen as the evalution metric for part segmentation and object categorization, respectively. The results are shown in the $2$nd and $3$rd column of Table~\ref{tab:partialseg}.

Our approach outperforms the baseline ACNN by a large margin on both segmentation and classification. Note that ACNN~\cite{boscaini2016learning} does not explicitly conduct multi-scale analysis and is designed for near-isometric 3D shapes with similar spectral domains, thus generalizes less well across a diverse set of shapes. Our framework, in contrast, could effectively capture multi-scale context information, a feature that is highly important for both segmentation and classification. The spectral domain synchronization ability of SpecTN  further improves our generalizability, leading to an extra performance gain as is shown in Table~\ref{tab:partialseg}.

\mypara{Partial data part segmentation}
To evaluate the robustness of our approach to incomplete data, we conduct part segmentation on simulated scans of 3D shapes from a single viewpoint. To be specific, we generate $N=6$ simulated scans for each 3D shape in the part annotation dataset \cite{Yi16} from random viewpoints, and then use these partial point cloud with part annotations for train and test. All the partial point clouds are normalized to fit into a unit cube. Following the train/test split provided by \cite{shapenet2015}, we train our network to segment shape parts for each category. Again we compare our method with ACNN~\cite{boscaini2016learning}. IoU is used as evaluation metric and the results are shown in the $4$th and $5$th column of Table~\ref{tab:partialseg}.

Our approach outperforms the baseline on partial data part segmentation by a large margin. In particular, from complete shape to partial shape setting, the performance drop of our approach is less significantly than the baseline, reflected by the gap of mean IoU between the complete data setting and the partial setting. It verifies that our method is more robust to data incompleteness. We surmise that the performance of ACNN is heavily influenced by noisy and sensitive principal curvature estimation on partial scans since this step plays a crucial rule in determining its local frames; whereas our approach makes less assumption about quality of the underlying shape.

\begin{table}[t!]
\centering
\small
\begin{tabular}{@{}c|cc|cc}
\hline
& cross cat IoU & Acc & partial & complete\\ \hline
ACNN & 69.22 & 93.99 & 69.21 & 79.63 \\ \hline
Ours1 & 79.65 & 99.59 & 76.19 & 83.48 \\
Ours2 & \textbf{81.97} & \textbf{99.71} & \textbf{78.02} & \textbf{84.74} \\ \hline
\end{tabular}
\caption{The $2$nd and $3$rd column of the table reports IoU for cross category part segmentation along with an induced classification accuracy. $4$th and $5$th column of the table reports IoU for part segmentation on partial shapes and complete shapes correspondingly. Our1 and Our2 corresponds to our framework without and with SpecTN respectively. In all experiments we beat the baseline by a large margin.}
\label{tab:partialseg}
\vspace{-0.6cm}
\end{table}

\subsection{Keypoint Prediction}
\begin{figure}[t!]
    \centering
    \includegraphics[width=0.8\linewidth]{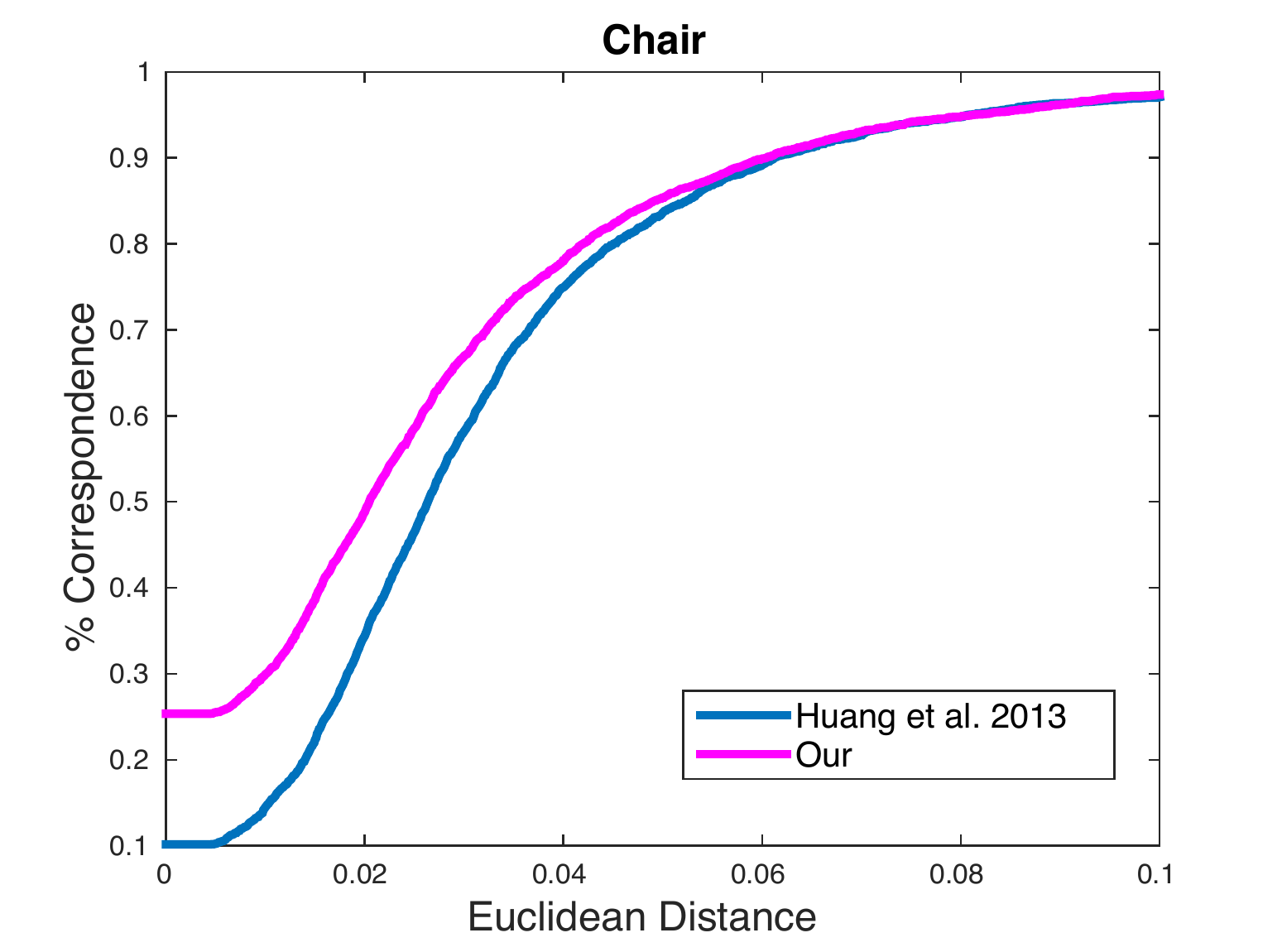}
    \caption{Keypoint prediction comparison. We draw PCK curves for both methods while changing the error threshold. Our approach outperforms \cite{huang2013fine} on average and has particularly high local accuracy when the error threshold is small, i.e. our approach reaches $pck=0.29$ when error threshold equals $0.01$, while \cite{huang2013fine} reaches $pck=0.16$}
    \label{fig:keypoint}
    \vspace{-0.5cm}
\end{figure}

Our framework is not limited to part segmentation but could learn more general functions on graphs. In this section, we evaluate our framework on the keypoint prediction task. We associate each keypoint an individual label and assign all the non-keypoints a background class label. The keypoint prediction problem could be treated as a multi-class classification problem and the cross-entropy loss is optimized during training. We evaluate our approach against previous state-of-the-art method \cite{huang2013fine}. \cite{huang2013fine} first jointly aligns all the shapes in 3D space via free-form deformation and then propagates keypoint labels to test shapes from its $K$ nearest training shapes. We manually tune $K$ and report the best performance of this method. Five-folds cross validation is adopted during evaluation, and PCK (percentage of correct keypoints) is used as evaluation metric. We show the PCK curve for the two approaches in Figure~\ref{fig:keypoint}. Each point on a curve indicates fraction of correctly predicted keypoints for a given Euclidean error threshold. Our approach outperforms \cite{huang2013fine}, in particular, more precise predictions can be obtained by our method (see the region close to y-axis).

\subsection{Normal Prediction}
\label{sec:normal}
\begin{figure}
 \centering
 \includegraphics[width=1\linewidth]{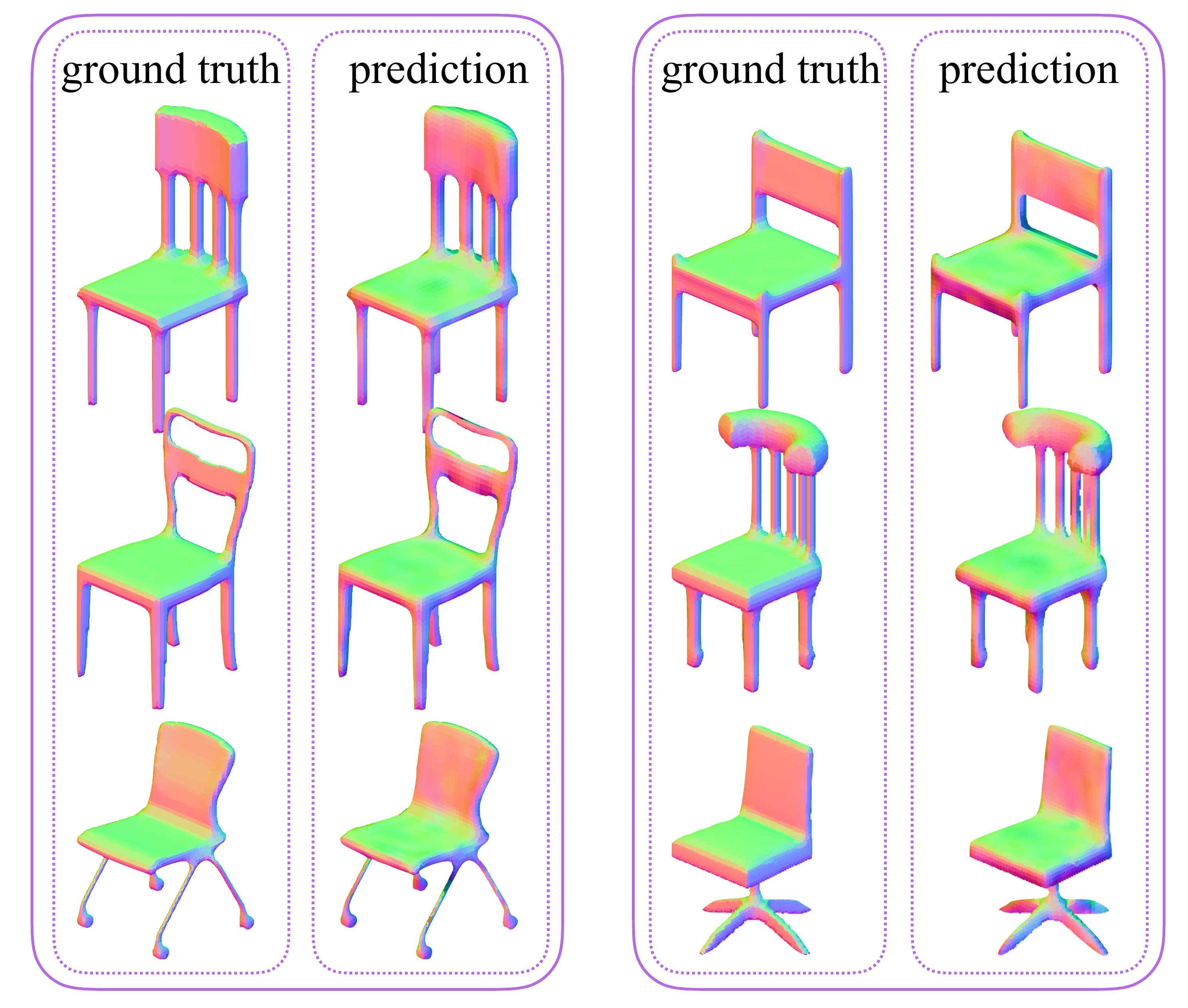}
 \caption{We evaluate our framework on normal prediction task. The colors shown on the 3D shape are RGB-coded normals, namely putting XYZ components of normal directions into RGB channels. Our framework could predict reasonable normal directions even on very thin structures.}
 \label{fig:normpred}
\end{figure}

To further validate the generality of our framework, we leverage our proposed SyncSpecCNN to learn another type of graph vertex function, vertex normal function. Specifically, our SyncSpecCNN takes the XYZ coordinate function of graph vertices as network input and predicts vertex normal as output. The network is trained to minimize the L2 loss between ground truth normals and predicted normals. We use the official train/test split provided by \cite{shapenet2015} and visualize some of the normal prediction results from test set in Figure~\ref{fig:normpred}.

It can be seen our predictions are very close to the ground truth at most of the time.Even on thin structures the normal predictions are still reasonable. One problem of our prediction is that it tends to generate smoothly transiting normals along the boundary while the ground truth is sharper. This is due to the fact that we are using a small number of eigenbases in our experiments, which is not friendly to regression tasks with very high frequency signal as target.

\subsection{Diagnosis}
\myparaly{Spectral Dilated Kernel Parametrization}
 We evaluate our dilated kernel parametrization from two aspects: the basis function choice and kernel scale choice. Table~\ref{tab:kerneldesign} summarizes all the comparison results, as explained below.
 
We explore the expressive power of different kernel basis. In the family of spectral CNN, convolution kernels are parametrized by a linear combination of basis functions, i.e. modulated exponential window in our case. Previous methods have proposed to use different basis functions such as cubic spline basis \cite{bruna2013spectral} and exponential window basis \cite{boscaini2016learning}. Each row of Table~\ref{tab:kerneldesign} corresponds to a basis choice.
 
We also evaluate the effectiveness of multi-scale analysis by changing the spatial sizes of convolution kernels. We compare with two baseline choices: set all kernel size to be the smallest kernel size in the current network; set to be the largest one. Each column of Table~\ref{tab:kerneldesign} corresponds to a kernel scale choice.
 
All numbers are reported on the cross-category part segmentation task, by IoU. We only take the XYZ coordinate function of graph vertices as network input as opposed to handcrafted geometry features which may have already capture some multi-scale information. Also we remove the $7$th and $8$th layers from our network which involves SpecTN and is designed for very large convolution kernels. 

It can be seen that modulated exponential window basis has a better expressive power compared with baselines for our segmentation task. Using multi-scale kernels also enables the aggregation of multi-scale information, thus producing better performance than small or large kernels alone. 

\begin{table}[h!]
\centering
\small
{
\begin{tabular}{@{}lccc}
\toprule
 & small & large & multiscale \\ \midrule
Cubic Spline & 0.5369 & \multicolumn{1}{c}{-} & \multicolumn{1}{c}{-} \\
Exp Window & 0.6285 & 0.7223 & 0.7386 \\
Modulated Exp Window & 0.6997 & 0.7341 & \textbf{0.7524} \\ \bottomrule
\end{tabular}
}
\caption{We compare different kernel basis and kernel size choices, using cross category part segmentation task for evaluation. IoU is reported in the table. In particular, we compare cubic spline basis \cite{bruna2013spectral}, exponential window basis \cite{boscaini2016learning} and our modulated exponential window. All convolution kernels are parametrized by the same number of parameters and we tweak the hyper parameters of different basis functions so that their spatial sizes are comparable. We also compare three different kernel size choices. "small" indicates using small convolution kernel only; "large" indicates using large convolution kernel only; "multiscale" uses kernels of different sizes in different layers, as in our current design. It's not obvious how to parametrize multi-scale convolution kernels using cubic spline basis functions, therefore we evaluate cubic spline basis with small-sized kernels only.} 
\label{tab:kerneldesign}
\vspace{-0.3cm}
\end{table}

\begin{figure}
 \centering
 \includegraphics[width=0.8\linewidth]{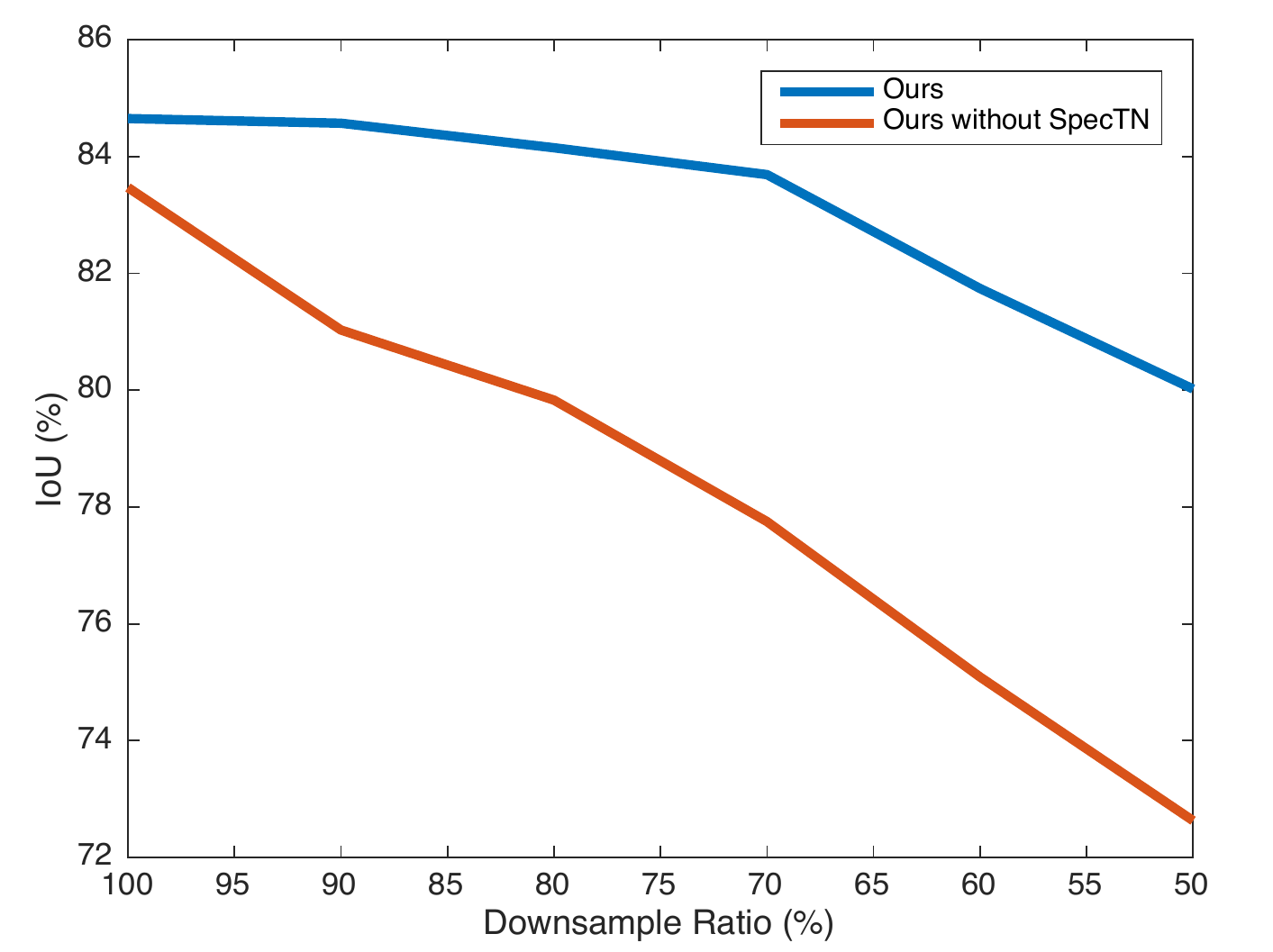}
 \caption{We evaluate the robustness of our model to sampling density change. Test shapes are downsampled by different ratios and fed into our network. We compute the segmentation IoU for different downsample ratios and show it here. With SpecTN, our framework becomes more robust to sampling density change.}
 \label{fig:downsample}
\end{figure}

\paragraph{Robustness to Sampling Density Variance}
In this experiment, we evaluate the robustness of our approach w.r.t point cloud density variation. To be specific, we train our SyncSpecCNN for shape segmentation on the point cloud provided by \cite{Yi16} first. Then we downsample the point cloud under different downsample ratio and evaluate our trained model to check how segmentation performance would change. Again we evaluate our approach with/without SpecTN and the result is shown in Figure~\ref{fig:downsample}.
 
By introducing SpecTN, our framework becomes more robust to sampling density variation. Our conjecture is that sampling density variation may result in large spectral space perturbation, therefore being able to synchronize different spectral domains becomes especially important.

\subsection{Qualitative Results and Error Analysis}
\begin{figure}
    \centering
    \includegraphics[width=\linewidth]{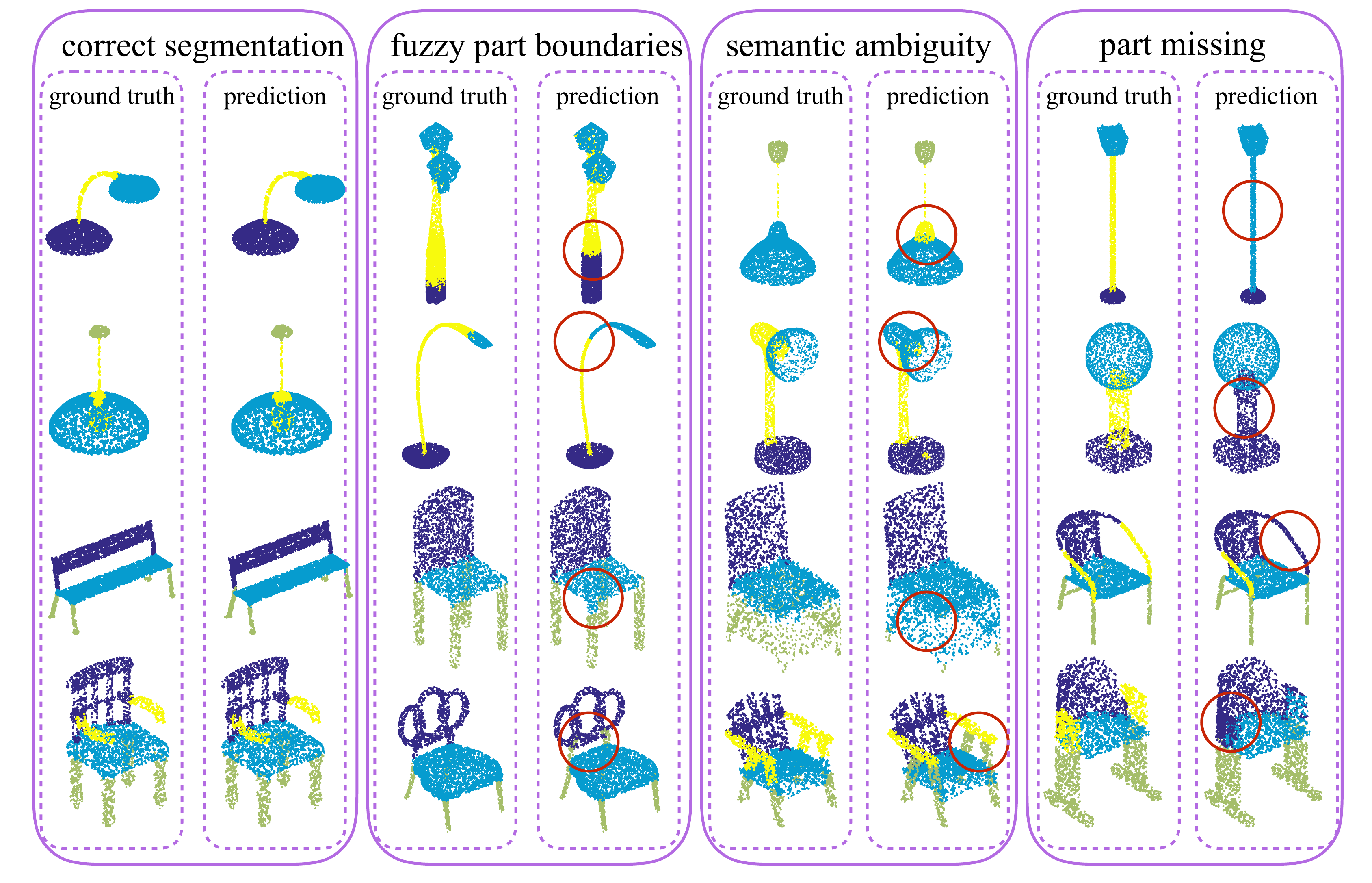}
    \caption{We visualize some segmentation results from our network prediction. The first block shows typical correct segmentations, notice the huge shape variation we can cover. The second to fourth blocks summarize different error patterns we observe in the results.}
    \label{fig:erroranalysis}
\end{figure}

Figure~\ref{fig:erroranalysis} shows segmentation results generated from our network on two categories, Chair and Lamp. Representative good results are shown in the first block and  typical error patterns are summarized from the second to fourth blocks.

Most of our segmentation is very close to ground truth as is shown in the first block. We can accurately segment shapes with large geometric or topological variations like wide bench v.s. ordinary chair, pendant lamp v.s. table lamp. The lamp base on the first row and the lampshade on the second row are very similar regarding their local geometry; however, since our network is able to capture large scale context information, it could still differentiate the two and segment shapes correctly.

We observe several typical error patterns in our results. Most segmentation error occurs along part boundaries. 
There are also cases where the semantic definition of parts has inherent ambiguities. 
We also observe a third type of error pattern, in which our prediction might miss a certain part completely, as is shown in the fourth block.

\section{Conclusion}
\label{sec:conclusion}
We introduce a novel neural network architecture, the Synchronized Spectral CNN (SyncSpecCNN), for semantic annotation on 3D shape graphs. To share coefficients and conduct multi-scale analysis in different parts of a single shape graph, we introduce a spectral parametrization of dilated convolutional kernels. To allow parameter sharing across related but different shapes that may be represented by very different graphs, we introduce a spectral transformer network to synchronize different spectral domains. The effectiveness of different components in our network is validated through extensive experiments. Jointly these contributions lead to state-of-the-art performance on various semantic annotation tasks including 3D shape part segmentation and 3D keypoint prediction.

{\small
\bibliographystyle{ieee}
\bibliography{acme}
}

\end{document}